\pdfoutput=1

\documentclass[11pt]{article}

\usepackage[]{acl}

\usepackage{times}
\usepackage{latexsym}

\usepackage[T1]{fontenc}

\usepackage[utf8]{inputenc}

\usepackage{microtype}

\usepackage{graphicx}       
\usepackage[T1]{fontenc}    
\usepackage{multirow}       
\usepackage{subcaption}     
\usepackage{bold-extra}     
\usepackage{bm}             
\usepackage[hang,flushmargin]{footmisc}  
\newcommand{\LN}{\linebreak\noindent}    
\usepackage{amsfonts,amsmath,amssymb}
\usepackage{array,booktabs}

\title{Online Coreference Resolution for Dialogue Processing:\\Improving Mention-Linking on Real-Time Conversations}

\author{
  Liyan Xu \quad\quad Jinho D. Choi \\
  Department of Computer Science \\
  Emory University, Atlanta, USA \\ \{\texttt{liyan.xu,jinho.choi}\}\texttt{@emory.edu}}

\begin{document}
\maketitle

\begin{abstract}

This paper suggests a direction of coreference resolution for online decoding on actively generated input such as dialogue, where the model accepts an utterance and its past context, then finds mentions in the current utterance as well as their referents, upon each dialogue turn.
A baseline and four incremental-updated models adapted from the mention-linking paradigm are proposed for this new setting, which address different aspects including the singletons, speaker-grounded encoding and cross-turn mention contextualization.
Our approach is assessed on three datasets: \textit{Friends}, \textit{OntoNotes}, and \textit{BOLT}.
Results show that each aspect brings out steady improvement, and our best models outperform the baseline by over 10\%, presenting an effective system for this setting.
Further analysis highlights the task characteristics, such as the significance of addressing the mention recall.


\end{abstract}
\section{Introduction}
\label{sec:introduction}

\begin{figure}[t]
\centering
\includegraphics[width=\columnwidth]{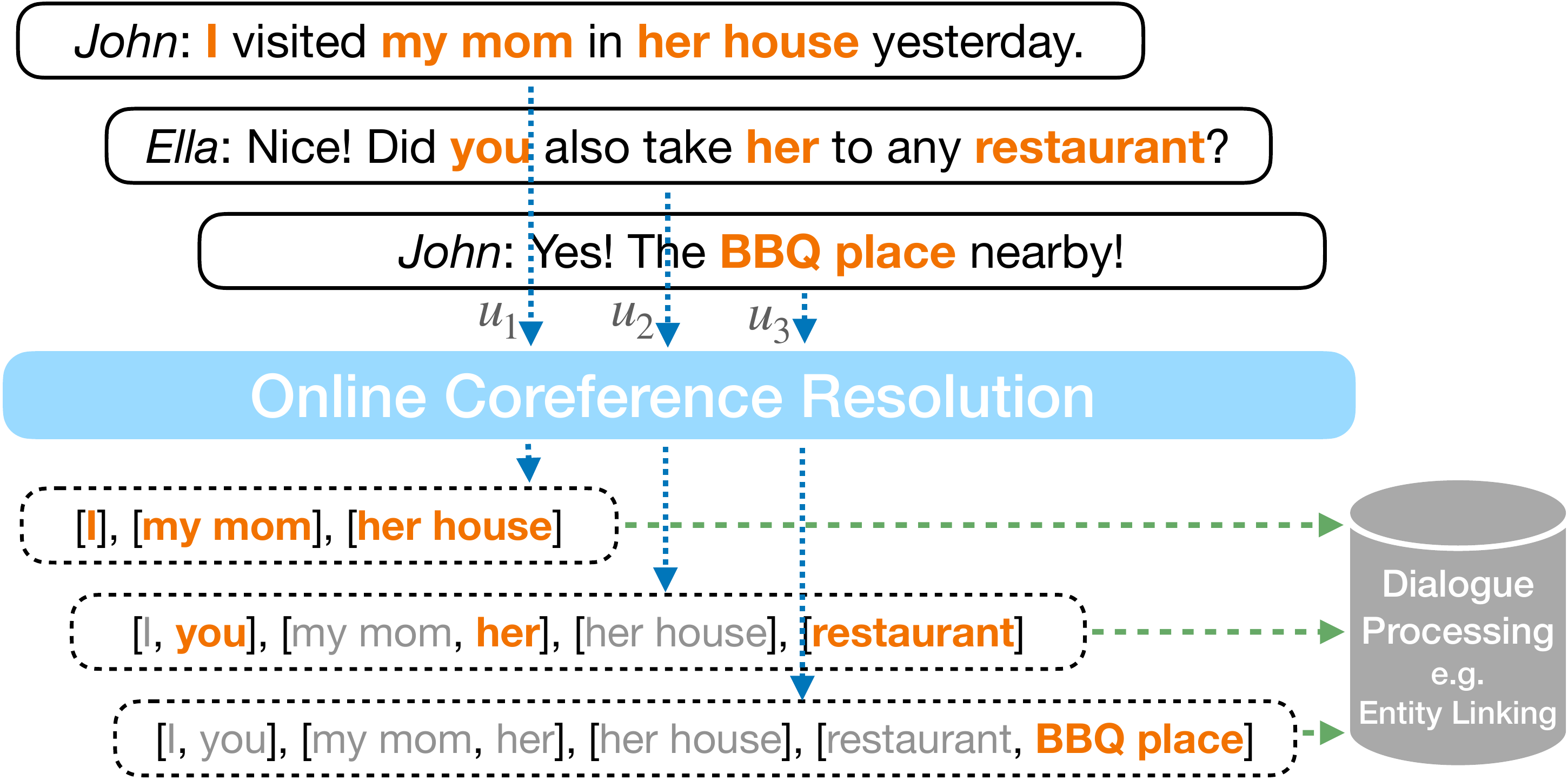}
\caption{Illustration of the online setting. Predictions upon each turn are made immediately and ready for consumption by downstream applications. New mentions at each turn are marked by boldface in orange.}
\label{fig:online}
\vspace{-2ex}
\end{figure}

It has been made practical recently to apply coreference resolution to assist a broad scope of NLP tasks \citep{peng-etal-2017-cross,sahu-etal-2019-inter,gao-etal-2019-interconnected}, especially with the advent of neural end-to-end decoding and contextualized encoding \citep{lee-etal-2017-end,lee-etal-2018-higher,joshi-etal-2019-bert,spanbert-joshi,wu-etal-2020-corefqa}.
However, it is quite limited to use existing coreference models in real-time dialogue processing systems, as most of them are not trained to handle an online decoding environment.
In the dialogue domain, recent efforts have focused on ellipsis recovery and query rewriting \citep{gecor,tseng-etal-2021-cread}; in this work, we target to address a new perspective specifically for the online decoding, where the model sequentially accepts utterances in a dialogue and spits out valid mentions as well as their referent links for each latest utterance turn upon arrival, to be consumed by the downstream dialogue processing (Figure~\ref{fig:online}).

More formally, let $u_i$ be the current ($i$'th) utterance in a dialogue ($u_1, .., u_i, ..$); $\mathcal{M}_{i}$ be the mentions in $u_i$; $\mathcal{M}^{i-1}$ be the mentions from previously predicted clusters till $u_{i-1}$. The objective upon $i$'th turn is to: (1) identify $\mathcal{M}_{i}$ (2) identify conference links among $\mathcal{M}_{i}$, as well as from $\mathcal{M}_{i}$ to $\mathcal{M}^{i-1}$. We do not allow updates on $\mathcal{M}_{i}$ later, since that would be equivalent to general coreference resolution; in this work, we specifically target this underexplored online scenario under this setting, which requires accurate predictions upon each turn that could be directly consumed by downstream applications.


Several quasi-online coreference models have been proposed that maintain and update referents sequentially \citep{clark-manning-2015-entity,clark-manning-2016-improving,liu-etal-2019-referential,toshniwal-etal-2020-learning,xia-etal-2020-incremental}.
However, these models
differ from our real online setting in two ways.
First, only the latest utterance and its past sequence are visible in our setting, so that decisions need to be made without knowing the unseen future.
Second, the decision of whether a span should be extracted or linked to others
needs to be made immediately at each utterance turn, while quasi-online models can maintain an internal pool of candidates and make one final prediction after the entire document is processed.

For this task, we first introduce our baseline adapted from the classic mention-linking (ML) approach \citep{wiseman-etal-2015-learning,lee-etal-2017-end}, and then propose four models where each one does an incremental update upon the previous model and addresses a specific perspective of this task, including the online inference, singletons, speaker-grounded encoding, and mention contextualization across utterance turns. For our approach, we do not use models that maintain explicit entities, because: (1) it has been shown that higher-order features from entity representation provide negative to marginal positive impact over ML counterparts despite their complexities \citep{xu-choi-2020-revealing,xia-etal-2020-incremental,toshniwal-etal-2020-learning}; (2) ML models are ``stateless'' so that they do not need to maintain decision states for previous mentions, which makes it more adaptable to applications in practice.


All models are evaluated on three datasets to test the generalizability of our approach, and the best model obtains over 10\% improvement over the baseline on all datasets. Results and further analysis suggest that each aforementioned aspect can bring out steady improvement under the online setting, and highlight the singleton recovery to be the most critical component.


\section{Approach}
\label{sec:approach}


\paragraph{End-to-End Resolution}

Our model backbone is based on the end-to-end coreference resolution \cite{lee-etal-2018-higher} with a Transformers encoder \cite{spanbert-joshi}.
It scores every span for being a mention, and extracts top spans as mention candidates. Pairwise scoring is then performed among all candidates to determine the coreference links. Details of the model architecture can be referred by the paper from \citet{lee-etal-2018-higher}, and we denote the original \textbf{c}oreference loss as $\mathcal{L}_c$.

\paragraph{Baseline (\texttt{BL})}

We first present our baseline that takes the end-to-end model and trains in the exact same non-online way as prior work, but adapts the decoding to fit in our online inference setting.

Let $u_i$ be the $i$'th utterance in the dialogue, and $|u_i|$ be its length (number of tokens).
During online decoding upon $u_i$, this model takes an utterance sequence with past context as input, denoted by $\mathcal{U}^i_k = (u_k, .., u_i)$; $k \in [1,i)$ is dynamically determined by $\sum_{j=k}^i |u_j| \leq \Upsilon$ where $\Upsilon$ is the max number of tokens that the encoder accepts. Different from \citet{lee-etal-2018-higher}, the mention candidates now consist of two parts: (1) the extracted top candidates solely from $u_i$, denoted as $\mathcal{X}_i$; (2) mentions from previously predicted clusters from $\mathcal{U}^{i-1}_k$, denoted as $\mathcal{M}_k^{i-1}$. Thereby the final candidate set $\mathcal{X}$ can be denoted as $\mathcal{X}_i \cup \mathcal{M}_k^{i-1}$. The same pairwise scoring as prior work is then performed on all candidates $\mathcal{X}$. Since we do not modify previous decisions in our setting, we keep coreference links among $\mathcal{X}_i$, or from $\mathcal{X}_i$ to $\mathcal{M}_k^{i-1}$, but not among $\mathcal{M}_k^{i-1}$. The predicted clusters after $u_i$ will be updated in the same way by picking the referent antecedents according to coreference links.

\paragraph{Singleton Recovery (\texttt{SR})}

\texttt{SR} is built upon \texttt{BL} to address the singleton problem. In \texttt{BL}, after processing each utterance sequence $\mathcal{U}^i_k$, the model filters out mention candidates from $\mathcal{X}_i$ that are not referent to any other candidates, according to the mention-linking paradigm.
However, it results on losing non-anaphoric mentions that do not have referents in $u_i$, and yields a critical issue for online inference because mentions in $u_i$ that are currently singletons but potentially will find referents in later utterances can get discarded too early.

To address this issue, we adopt a simple strategy similar to \citep{xu-choi-2021-adapted} that preserves any candidates whose mention scores are larger than a threshold of $0$, denoted as $s_m > 0$, and creates a singleton cluster for each of which have not yet found any referent (intermediate singletons).
However, as many annotation schemes do not require annotating singletons, e.g. CoNLL 2012, we may not have ``true'' gold labels covering every valid mentions, similar to the ``misguidance of unlabeled entities'' problem in named entity recognition (NER) \citep{li2021empirical}.
Let $\Psi^+_m$ be the set of $s_m$ of gold candidates according to the annotation, and $\Psi^-_m$ be the set of $s_m$ of other candidates that may also contain certain valid mentions (singletons).
We mitigate the false negative issue of unlabeled mentions by applying dynamic negative sampling on $\Psi^-_m$, denoted as $\Phi^-_m$, where $|\Psi^+_m| \approx |\Phi^-_m|$.
Binary cross-entropy (\texttt{BCE}) loss is then used for this optimization to aid the threshold requirement:
\begin{align}
\label{eq:mention_loss}
    \mathcal{L}_{m} &= \text{\texttt{BCE}} ( \Psi^+_m, \Phi^-_m ) \\
    \mathcal{L} &= \alpha_c \cdot \mathcal{L}_c + \alpha_m \cdot \mathcal{L}_m
\end{align}
The final loss $\mathcal{L}$ is estimated by the weighted sum of\LN $\mathcal{L}_m$ and $\mathcal{L}_c$ using the hyperparameters $\alpha_c$ and $\alpha_m$.



\paragraph{Online Resolution (\texttt{OR})}

\texttt{OR} is designed specifically for online inference on dialogues.
Distinguished from \texttt{BL} that takes the whole document as input in training, \texttt{OR} takes $\mathcal{U}^i_k$ as input for both training and decoding, closing the gap.
To capture subtle nuances from different speakers in the dialogue, we collect speaker names within each dialogue and assign a special token of position-based ID to each speaker (e.g. $\texttt{S}_1$, $\texttt{S}_2$) based on speaking orders,
which is then prepended to its corresponding utterance \cite{wu-etal-2020-corefqa}. We also add \texttt{[SEP]} before $u_i$ to signal the latest utterance. The following sequence is used as input for \texttt{OR}:
\begin{align}
\label{eq:or}
    \{\texttt{S}_k\}^\frown u_k^\frown \cdots^\frown\{\texttt{[SEP]}\}^\frown \{\texttt{S}_i\}^\frown u_i 
\end{align}
During training upon the $i$'th turn, gold mentions in $\mathcal{U}^{i-1}_k$ are used as $\mathcal{M}_k^{i-1}$; the losses $\mathcal{L}_m$ and $\mathcal{L}_c$ are estimated only on candidates from $u_i$. Gradient accumulation is applied across multiple utterance turns, and we warm-start \texttt{OR} by initializing from the parameters of \texttt{SR}, followed by the online training described above. The decoding step for \texttt{OR} is kept the same as \texttt{BL} and \texttt{SR}.

\paragraph{Speaker-Grounding (\texttt{SG})}

\texttt{SG} adds a speaker-grounding subtask upon \texttt{OR}, which is to facilitate the encoding of multi-speaker interaction which is an important aspect in dialogues.
In \texttt{OR}, although each input token is conditioned on speaker tokens as in Eq~\eqref{eq:or}, it is not obvious to the model that each token is from which speaker, which can be a barrier to learn the speaker interaction.
To explicitly regularize the speaker encoding, we add a subtask to predict whether two candidates are from the same speaker based on their embeddings: the model gives a same-speaker score $s_s$ such that pairs from the same speaker have $s_s > 0$ and others $s_s \leq 0$, forcing the semantic representation to fuse the speaker interaction.
Let $\Psi^+_s$ be the set of $s_s$ of pairs from the same speaker; $\Psi^-_s$ be the set of $s_s$ of other pairs.
We optimize $s_s$ by \texttt{BCE}, adding the loss in addition to $\mathcal{L}_c$ and $\mathcal{L}_m$:
\begin{align}
    s_{s}(x,y) &= w_s \cdot [g_x \oplus g_y  \oplus (g_x \circ g_y)  \oplus (g_x - g_y)] \nonumber\\
    \mathcal{L}_{s} &= \text{\texttt{BCE}} ( \Psi^+_s, \Psi^-_s ) \label{eq:speaker_loss} \\
    \mathcal{L} &= \alpha_c \cdot \mathcal{L}_c + \alpha_m \cdot \mathcal{L}_m + \alpha_s \cdot \mathcal{L}_s \label{eq:final_loss}
\end{align}
$g_x/g_y$ denotes the representation of a candidate and $w_s$ is the scoring parameter. $\oplus$ denotes concatenation and $\circ$ is the element-wise multiplication. 
We also apply negative sampling to keep $ |\Psi^+_s| \approx  |\Psi^-_s|$.


\paragraph{Span-Level Self-Attention (\texttt{SA}) }
\texttt{SA} is also added upon \texttt{OR} to achieve candidate contextualization.
For each input $\mathcal{U}^i_k$, the representation of all candidates $\mathcal{X}$ is contextualized on the token-level because of Transformers' encoding. However, $\mathcal{M}_k^{i-1}$ is not used until the pairwise scoring. Therefore, $\mathcal{X}_i$ is not explicitly conditioned on the previously extracted mentions ($\mathcal{M}_k^{i-1}$) on the span-level. To capture the dependency among all mention candidates across utterances, we pass $\mathcal{X}$ to a scaled dot-product self-attention layer \citep{NIPS2017_3f5ee243} before the pairwise scoring:
\begin{align}
\label{eq:self-attn}
    G' = \text{softmax} \big( \frac{(G W_q)(G W_k)^T}{\sqrt{d}} \big) (G W_v),
\end{align}
where $G \in \mathbb{R}^{|\mathcal{X}| \times d}$ is the embedding matrix of all candidates, $d$ is the embedding size, $W_q, W_k, W_v$ are the parameters. $G'$ is the new candidate-aware embedding matrix, which provides enhanced candidate representation for the pairwise scoring.

\section{Experiments}
\label{sec:experiments}

\begin{table*}[htbp!]
\centering
\resizebox{\textwidth}{!}{
\begin{tabular}{lcccc|cccc|cccc}
\toprule
 & \multicolumn{4}{c|}{\it Friends} & \multicolumn{4}{c|}{\it Onto-Conv} & \multicolumn{4}{c}{\it BOLT} \\
 \cmidrule{2-5} \cmidrule{6-9} \cmidrule{10-13}
 & MUC & B\textsuperscript{3} & CEAF\textsubscript{$\phi_4$} & Avg F1 & MUC & B\textsuperscript{3} & CEAF\textsubscript{$\phi_4$} & Avg F1 & MUC & B\textsuperscript{3} & CEAF\textsubscript{$\phi_4$} & Avg F1 \\
 \midrule
 \tt BL & 81.9 & 62.2 & 54.5 & 66.2 ($\pm$ 0.7) & 70.5 & 54.8 & 43.9 & 56.4 ($\pm$ 0.2) & 73.3 & 61.2 & 51.1 & 61.9 ($\pm$ 0.3) \\
 \tt SR & 85.5 & 68.3 & 61.7 & 71.8 ($\pm$ 0.5) & 77.5 & 63.2 & 55.2 & 65.2 ($\pm$ 0.6) & 79.6 & 71.8 & 61.7 & 71.0 ($\pm$ 0.4) \\
 \tt OR & 85.8 & 71.9 & 65.7 & 74.5 ($\pm$ 0.5) & 78.0 & 63.6 & 55.6 & 65.7 ($\pm$ 0.3) & 79.5 & 72.0 & 63.2 & 71.5 ($\pm$ 0.3) \\
 \; \tt +SG & 85.7 & 73.6 & 67.0 & 75.3 ($\pm$ 0.4) & 78.1 & \bf 64.3 & 56.5 & 66.3 ($\pm$ 0.3) & \bf 79.9 & 72.3 & 63.4 & 71.8 ($\pm$ 0.3) \\
 \; \tt +SG+SA & \bf 86.4 & \bf 73.7 & \bf 68.2 &  \textbf{76.1} ($\pm$ 0.1) & \bf 78.9 & \bf 64.3 & \bf 56.9 & \textbf{66.8} ($\pm$ 0.1) & \bf 79.9 & \bf 72.7 & \bf 64.1 & \textbf{72.3} ($\pm$ 0.2) \\
\bottomrule
\end{tabular}}
\caption{Results of all models in Section~\ref{sec:approach} on the evaluation sets of \textit{Friends}, \textit{Onto-Conv}, and \textit{BOLT} datasets. MUC, B\textsuperscript{3}, and CEAF\textsubscript{$\phi_4$} show the F1 scores of the corresponding metrics, and their macro-average score (Avg F1) is\LN used as the main evaluation metric. All scores presented here are the averaged scores over 3 repeated experiments; the standard deviations of Avg F1 scores are provided in the parentheses.}
\label{table:results}
\end{table*}

\paragraph{Datasets}
All models are experimented on the following three datasets.
\textit{Friends} contains transcripts from the TV show in which personal mentions are annotated for entity linking.
Each scene is considered an independent dialogue where utterances and speaker IDs are provided.
We adapt the data split suggested by \citet{zhou-choi-2018-exist}.
\textit{Onto-Conv} consists of documents in three genres selected from OntoNotes 5.0: broadcasting and telephone conversations, and web text including discussion forums.
We adapt the data split provided by \citet{pradhan-etal-2012-conll} and treat each document as a dialogue and every sentence as an utterance.
\textit{BOLT} follows the same annotation guideline as OntoNotes although documents are from discussion forums, SNS chats, and telephone conversations \citep{li-etal-2016-large}.
Since this is the first work using \textit{BOLT} for this task, we create a new data split for future replicability (see \ref{appx:dataset}).
Out of these three datasets, only \textit{Friends} provides annotation of singletons.

The numbers of documents in the training, development, and test set of \textit{Friends}, \textit{Onto-Conv}, \textit{BOLT} are provided in Table~\ref{table:dataset_stats}, along with the averaged numbers of speakers, entity clusters and utterances per document of each dataset. More details regarding the datasets are provided in Appendix \ref{appx:dataset}.

\begin{table}[htbp!]
\small
\centering
\begin{tabular}{c|ccc|ccc}
\multicolumn{1}{c|}{} & \multicolumn{1}{c}{TRN} & \multicolumn{1}{c}{DEV} & \multicolumn{1}{c}{TST} & \multicolumn{1}{|c}{NS} & \multicolumn{1}{c}{NC} & \multicolumn{1}{c}{NU} \\
\midrule
\texttt{F} & 987 & 122 & 192 & 3.7 & 4.6 & 18.7 \\
\texttt{O} & 566 & 100 & 95 & 2.4 & 16.2 & 49.5 \\
\texttt{B} & 943 & 117 & 117 & 2.9 & 9.2 & 18.1 \\
\end{tabular}
\caption{Statistics of the dataset \textit{Friends} (\texttt{F}), \textit{Onto-Conv} (\texttt{O}), \textit{BOLT} (\texttt{B}). TRN, DEV, TST are the numbers of documents in the training, development, and test set of each dataset. NS, NC, NU are the averaged numbers of speakers, entity clusters, utterances per document of each dataset.}
\label{table:dataset_stats}
\end{table}

\paragraph{Settings}


Our implementation are based on the PyTorch coreference models from \citet{xu-choi-2020-revealing}, and SpanBERT\textsubscript{BASE} is adopted as the encoder. The implementation and trained models have been partially integrated with the open source project ELIT\footnote{\url{https://github.com/emorynlp/elit}} \citep{he2021elit}.

During inference, all predicted clusters are collected and merged accordingly across utterances, and get evaluated by comparing them to the ground truth (all gold non-singleton clusters) at the end of each dialogue, in the same way as the CoNLL'12 shared task protocol. Detailed experimental settings are provided in Appendix \ref{appx:implementation}.

\paragraph{Results}

Table~\ref{table:results} describes the performance of all models on the test sets in the three datasets. 
These results are averaged across 3 repeated experiments; Avg-F1 is used as the main evaluation metric.
Each proposed model gives steady improvement, and the best result is achieved by the \texttt{OR+SG+SA} model, surpassing the \texttt{BL} model on all datasets by significant margins of $\approx$10\%.
Among these models, singleton recovery contributes the most upon \texttt{BL}, demonstrating that albeit simple and intuitive, the training and inference of intermediate singletons is essential in online coreference resolution. 

\subsection{Analysis on Online Inference}
\label{subsec:analysis_online}

To identify how model predictions are affected by online inference, all mentions in the predicted clusters are examined against the gold clusters.
Table~\ref{table:analysis_online} shows the results of mention precision and recall from the four experimental settings. 

\begin{table}[htbp!]
\centering
\resizebox{0.91\columnwidth}{!}{
\begin{tabular}{lccccccccccc}
& \multicolumn{2}{c}{\it Friends} && \multicolumn{2}{c}{\it Onto-Conv} && \multicolumn{2}{c}{\it BOLT} \\
\cmidrule{2-3} \cmidrule{5-6}  \cmidrule{8-9}
& P & R && P & R && P & R \\
\midrule
 \tt N:BL & 92.0 & 92.5 && 88.1 & \bf 83.6 && 85.2 & \bf 82.8 \\
 \tt O:BL & \bf 92.5 & 85.3 && \bf 92.1 & 60.6 && \bf 89.0 & 64.8 \\
 \tt O:SR & \bf 92.5 & \bf 93.2 && 89.4 & 78.8 && 87.4 & 78.3 \\
 \tt O:SR- & \bf 92.5 & 92.5 && 90.4 & 74.8 && 88.4 & 76.7 \\
\end{tabular}}
\caption{The \textbf{P}recision and \textbf{R}ecall of all mentions in the predicted clusters on the test sets in the three datasets.\LN \texttt{N} is \textbf{N}on-online inference as in CoNLL'12 shared task, \texttt{O} is \textbf{O}nline inference as in this work. \texttt{SR-} is the \textbf{S}ingleton \textbf{R}ecovery (\texttt{SR}) model without applying negative sampling on the mention loss in training.}
\label{table:analysis_online}
\end{table}

\noindent Following observations are drawn by this analysis:

\noindent \textbf{(1)} Comparing \texttt{N:BL} and \texttt{O:BL}, online inference indeed leads to a large drop on the mention recall as expected, without as much increase on precision, due to the omission of intermediate singletons.

\noindent \textbf{(2)} Comparing \texttt{O:BL} and \texttt{O:SR}, singleton recovery (\texttt{SR}) significantly improves the mention recall (8\% for \textit{Friends} and 13+\% for others) without sacrificing much precision.
However, notice that the recall of \texttt{O:SR} for \textit{Friends} is even higher than that of non-online inference (\texttt{N:BL}), but the recall for \textit{Onto-Conv} and \textit{BOLT} is still 4+\% lower than that of \texttt{N:BL}.
This is due to the fact that \textit{Friends} does have singletons annotated while the other two do not. 
Thus, \texttt{O:SR} for \textit{Friends} does not suffer from the ``misguidance of unlabeled entities'' problem.

\noindent \textbf{(3)} Comparing \texttt{O:SR} and \texttt{O:SR-}, it illustrates the positive impact of applying negative sampling on mentions to alleviate the false-negative issue of unlabeled mentions, which improves recall while maintaining similar precision for online inference.

\subsection{Analysis on Utterance Interaction}
\label{subsec:analysis_speaker}

As we aim to build a robust online resolution model in the dialogue domain, understanding of individual speakers is important especially in multi-party interaction.
In comparison to the binary indicator used in \texttt{BL} and \texttt{SR} that can handle only up to two speakers, adding the subtask for speaker-grounded encoding is shown to perform better for multi-speaker dialogues:
the improvement of \texttt{OR+SG} over \texttt{SR} is 3.5\% F1 for \textit{Friends}, but around 1\% F1 for the other two.
Our statistics show that 43\% dialogues in \textit{Friends} have at least 4 speakers, while being only 15\% and 24\% for the other two, suggesting that the multi-speaker environment indeed benefits more from the new speaker encoding scheme.

In addition, the percentages of pronouns in the gold mentions are 80.3\%, 53.5\%, and 63.5\% in \textit{Friends}, \textit{Onto-Conv}, and \textit{BOLT} respectively, which also highlights the importance of a better encoding scheme to handle a large portion of pronouns present in dialogue.
Thus, we suggest to employ a more advanced dialogue encoding that utilizes the speaker interaction clues as one of the future research direction for this online-decoding task.

\section{Conclusion}
\label{sec:conclusion}

This paper presents a new coreference resolution direction that aims towards an online decoding setting for dialogue processing.
A baseline and four incremental-updated models are proposed and evaluated on three datasets of the dialogue domain, and the best-performing model shows significant improvement over the baseline by $\approx$10\% F1. Further analysis suggests the importance of mention recall and speaker encoding, which could serve as the next future directions of this online setting.


\bibliography{references}

\begin{thebibliography}{24}
\expandafter\ifx\csname natexlab\endcsname\relax\def\natexlab#1{#1}\fi

\bibitem[{Clark and Manning(2015)}]{clark-manning-2015-entity}
Kevin Clark and Christopher~D. Manning. 2015.
\newblock \href {https://doi.org/10.3115/v1/P15-1136} {Entity-centric
  coreference resolution with model stacking}.
\newblock In \emph{Proceedings of the 53rd Annual Meeting of the Association
  for Computational Linguistics and the 7th International Joint Conference on
  Natural Language Processing (Volume 1: Long Papers)}, pages 1405--1415,
  Beijing, China. Association for Computational Linguistics.

\bibitem[{Clark and Manning(2016)}]{clark-manning-2016-improving}
Kevin Clark and Christopher~D. Manning. 2016.
\newblock \href {https://doi.org/10.18653/v1/P16-1061} {Improving coreference
  resolution by learning entity-level distributed representations}.
\newblock In \emph{Proceedings of the 54th Annual Meeting of the Association
  for Computational Linguistics (Volume 1: Long Papers)}, pages 643--653,
  Berlin, Germany. Association for Computational Linguistics.

\bibitem[{Gao et~al.(2019)Gao, Li, King, and
  Lyu}]{gao-etal-2019-interconnected}
Yifan Gao, Piji Li, Irwin King, and Michael~R. Lyu. 2019.
\newblock \href {https://doi.org/10.18653/v1/P19-1480} {Interconnected question
  generation with coreference alignment and conversation flow modeling}.
\newblock In \emph{Proceedings of the 57th Annual Meeting of the Association
  for Computational Linguistics}, pages 4853--4862, Florence, Italy.
  Association for Computational Linguistics.

\bibitem[{He et~al.(2021)He, Xu, and Choi}]{he2021elit}
Han He, Liyan Xu, and Jinho~D. Choi. 2021.
\newblock \href {http://arxiv.org/abs/2109.03903} {Elit: Emory language and
  information toolkit}.

\bibitem[{Joshi et~al.(2020)Joshi, Chen, Liu, Weld, Zettlemoyer, and
  Levy}]{spanbert-joshi}
Mandar Joshi, Danqi Chen, Yinhan Liu, Daniel~S. Weld, Luke Zettlemoyer, and
  Omer Levy. 2020.
\newblock \href {https://doi.org/10.1162/tacl\_a\_00300} {Spanbert: Improving
  pre-training by representing and predicting spans}.
\newblock \emph{Transactions of the Association for Computational Linguistics},
  8:64--77.

\bibitem[{Joshi et~al.(2019)Joshi, Levy, Zettlemoyer, and
  Weld}]{joshi-etal-2019-bert}
Mandar Joshi, Omer Levy, Luke Zettlemoyer, and Daniel Weld. 2019.
\newblock \href {https://doi.org/10.18653/v1/D19-1588} {{BERT} for coreference
  resolution: Baselines and analysis}.
\newblock In \emph{Proceedings of the 2019 Conference on Empirical Methods in
  Natural Language Processing and the 9th International Joint Conference on
  Natural Language Processing (EMNLP-IJCNLP)}, pages 5803--5808, Hong Kong,
  China. Association for Computational Linguistics.

\bibitem[{Lee et~al.(2017)Lee, He, Lewis, and Zettlemoyer}]{lee-etal-2017-end}
Kenton Lee, Luheng He, Mike Lewis, and Luke Zettlemoyer. 2017.
\newblock \href {https://doi.org/10.18653/v1/D17-1018} {End-to-end neural
  coreference resolution}.
\newblock In \emph{Proceedings of the 2017 Conference on Empirical Methods in
  Natural Language Processing}, pages 188--197, Copenhagen, Denmark.
  Association for Computational Linguistics.

\bibitem[{Lee et~al.(2018)Lee, He, and Zettlemoyer}]{lee-etal-2018-higher}
Kenton Lee, Luheng He, and Luke Zettlemoyer. 2018.
\newblock \href {https://doi.org/10.18653/v1/N18-2108} {Higher-order
  coreference resolution with coarse-to-fine inference}.
\newblock In \emph{Proceedings of the 2018 Conference of the North {A}merican
  Chapter of the Association for Computational Linguistics: Human Language
  Technologies, Volume 2 (Short Papers)}, pages 687--692, New Orleans,
  Louisiana. Association for Computational Linguistics.

\bibitem[{Li et~al.(2016)Li, Palmer, Xue, Ramshaw, Maamouri, Bies, Conger,
  Grimes, and Strassel}]{li-etal-2016-large}
Xuansong Li, Martha Palmer, Nianwen Xue, Lance Ramshaw, Mohamed Maamouri, Ann
  Bies, Kathryn Conger, Stephen Grimes, and Stephanie Strassel. 2016.
\newblock \href {https://www.aclweb.org/anthology/L16-1145} {Large
  multi-lingual, multi-level and multi-genre annotation corpus}.
\newblock In \emph{Proceedings of the Tenth International Conference on
  Language Resources and Evaluation ({LREC}'16)}, pages 906--913,
  Portoro{\v{z}}, Slovenia. European Language Resources Association (ELRA).

\bibitem[{Li et~al.(2021)Li, lemao liu, and Shi}]{li2021empirical}
Yangming Li, lemao liu, and Shuming Shi. 2021.
\newblock \href {https://openreview.net/forum?id=5jRVa89sZk} {Empirical
  analysis of unlabeled entity problem in named entity recognition}.
\newblock In \emph{International Conference on Learning Representations}.

\bibitem[{Liu et~al.(2019)Liu, Zettlemoyer, and
  Eisenstein}]{liu-etal-2019-referential}
Fei Liu, Luke Zettlemoyer, and Jacob Eisenstein. 2019.
\newblock \href {https://doi.org/10.18653/v1/P19-1593} {The referential reader:
  A recurrent entity network for anaphora resolution}.
\newblock In \emph{Proceedings of the 57th Annual Meeting of the Association
  for Computational Linguistics}, pages 5918--5925, Florence, Italy.
  Association for Computational Linguistics.

\bibitem[{Peng et~al.(2017)Peng, Poon, Quirk, Toutanova, and
  Yih}]{peng-etal-2017-cross}
Nanyun Peng, Hoifung Poon, Chris Quirk, Kristina Toutanova, and Wen-tau Yih.
  2017.
\newblock \href {https://doi.org/10.1162/tacl_a_00049} {Cross-sentence n-ary
  relation extraction with graph {LSTM}s}.
\newblock \emph{Transactions of the Association for Computational Linguistics},
  5:101--115.

\bibitem[{Pradhan et~al.(2012)Pradhan, Moschitti, Xue, Uryupina, and
  Zhang}]{pradhan-etal-2012-conll}
Sameer Pradhan, Alessandro Moschitti, Nianwen Xue, Olga Uryupina, and Yuchen
  Zhang. 2012.
\newblock \href {https://www.aclweb.org/anthology/W12-4501} {{C}o{NLL}-2012
  shared task: Modeling multilingual unrestricted coreference in
  {O}nto{N}otes}.
\newblock In \emph{Joint Conference on {EMNLP} and {C}o{NLL} - Shared Task},
  pages 1--40, Jeju Island, Korea. Association for Computational Linguistics.

\bibitem[{Quan et~al.(2019)Quan, Xiong, Webber, and Hu}]{gecor}
Jun Quan, Deyi Xiong, Bonnie Webber, and Changjian Hu. 2019.
\newblock \href {https://doi.org/10.18653/v1/D19-1462} {{GECOR}: An end-to-end
  generative ellipsis and co-reference resolution model for task-oriented
  dialogue}.
\newblock In \emph{Proceedings of the 2019 Conference on Empirical Methods in
  Natural Language Processing and the 9th International Joint Conference on
  Natural Language Processing (EMNLP-IJCNLP)}, pages 4547--4557, Hong Kong,
  China. Association for Computational Linguistics.

\bibitem[{Sahu et~al.(2019)Sahu, Christopoulou, Miwa, and
  Ananiadou}]{sahu-etal-2019-inter}
Sunil~Kumar Sahu, Fenia Christopoulou, Makoto Miwa, and Sophia Ananiadou. 2019.
\newblock \href {https://doi.org/10.18653/v1/P19-1423} {Inter-sentence relation
  extraction with document-level graph convolutional neural network}.
\newblock In \emph{Proceedings of the 57th Annual Meeting of the Association
  for Computational Linguistics}, pages 4309--4316, Florence, Italy.
  Association for Computational Linguistics.

\bibitem[{Toshniwal et~al.(2020)Toshniwal, Wiseman, Ettinger, Livescu, and
  Gimpel}]{toshniwal-etal-2020-learning}
Shubham Toshniwal, Sam Wiseman, Allyson Ettinger, Karen Livescu, and Kevin
  Gimpel. 2020.
\newblock \href {https://www.aclweb.org/anthology/2020.emnlp-main.685}
  {Learning to {I}gnore: {L}ong {D}ocument {C}oreference with {B}ounded
  {M}emory {N}eural {N}etworks}.
\newblock In \emph{Proceedings of the 2020 Conference on Empirical Methods in
  Natural Language Processing (EMNLP)}, pages 8519--8526, Online. Association
  for Computational Linguistics.

\bibitem[{Tseng et~al.(2021)Tseng, Bhargava, Lu, Moniz, Piraviperumal, Li, and
  Yu}]{tseng-etal-2021-cread}
Bo-Hsiang Tseng, Shruti Bhargava, Jiarui Lu, Joel Ruben~Antony Moniz, Dhivya
  Piraviperumal, Lin Li, and Hong Yu. 2021.
\newblock \href {https://doi.org/10.18653/v1/2021.naacl-main.265} {{CREAD}:
  Combined resolution of ellipses and anaphora in dialogues}.
\newblock In \emph{Proceedings of the 2021 Conference of the North American
  Chapter of the Association for Computational Linguistics: Human Language
  Technologies}, pages 3390--3406, Online. Association for Computational
  Linguistics.

\bibitem[{Vaswani et~al.(2017)Vaswani, Shazeer, Parmar, Uszkoreit, Jones,
  Gomez, Kaiser, and Polosukhin}]{NIPS2017_3f5ee243}
Ashish Vaswani, Noam Shazeer, Niki Parmar, Jakob Uszkoreit, Llion Jones,
  Aidan~N Gomez, \L~ukasz Kaiser, and Illia Polosukhin. 2017.
\newblock \href
  {https://proceedings.neurips.cc/paper/2017/file/3f5ee243547dee91fbd053c1c4a845aa-Paper.pdf}
  {Attention is all you need}.
\newblock In \emph{Advances in Neural Information Processing Systems},
  volume~30, pages 5998--6008. Curran Associates, Inc.

\bibitem[{Wiseman et~al.(2015)Wiseman, Rush, Shieber, and
  Weston}]{wiseman-etal-2015-learning}
Sam Wiseman, Alexander~M. Rush, Stuart Shieber, and Jason Weston. 2015.
\newblock \href {https://doi.org/10.3115/v1/P15-1137} {Learning anaphoricity
  and antecedent ranking features for coreference resolution}.
\newblock In \emph{Proceedings of the 53rd Annual Meeting of the Association
  for Computational Linguistics and the 7th International Joint Conference on
  Natural Language Processing (Volume 1: Long Papers)}, pages 1416--1426,
  Beijing, China. Association for Computational Linguistics.

\bibitem[{Wu et~al.(2020)Wu, Wang, Yuan, Wu, and Li}]{wu-etal-2020-corefqa}
Wei Wu, Fei Wang, Arianna Yuan, Fei Wu, and Jiwei Li. 2020.
\newblock \href {https://doi.org/10.18653/v1/2020.acl-main.622} {{C}oref{QA}:
  Coreference resolution as query-based span prediction}.
\newblock In \emph{Proceedings of the 58th Annual Meeting of the Association
  for Computational Linguistics}, pages 6953--6963, Online. Association for
  Computational Linguistics.

\bibitem[{Xia et~al.(2020)Xia, Sedoc, and
  Van~Durme}]{xia-etal-2020-incremental}
Patrick Xia, Jo{\~a}o Sedoc, and Benjamin Van~Durme. 2020.
\newblock \href {https://www.aclweb.org/anthology/2020.emnlp-main.695}
  {Incremental neural coreference resolution in constant memory}.
\newblock In \emph{Proceedings of the 2020 Conference on Empirical Methods in
  Natural Language Processing (EMNLP)}, pages 8617--8624, Online. Association
  for Computational Linguistics.

\bibitem[{Xu and Choi(2020)}]{xu-choi-2020-revealing}
Liyan Xu and Jinho~D. Choi. 2020.
\newblock \href {https://www.aclweb.org/anthology/2020.emnlp-main.686}
  {Revealing the myth of higher-order inference in coreference resolution}.
\newblock In \emph{Proceedings of the 2020 Conference on Empirical Methods in
  Natural Language Processing (EMNLP)}, pages 8527--8533, Online. Association
  for Computational Linguistics.

\bibitem[{Xu and Choi(2021)}]{xu-choi-2021-adapted}
Liyan Xu and Jinho~D. Choi. 2021.
\newblock \href {https://doi.org/10.18653/v1/2021.codi-sharedtask.6} {Adapted
  end-to-end coreference resolution system for anaphoric identities in
  dialogues}.
\newblock In \emph{Proceedings of the CODI-CRAC 2021 Shared Task on Anaphora,
  Bridging, and Discourse Deixis in Dialogue}, pages 55--62, Punta Cana,
  Dominican Republic. Association for Computational Linguistics.

\bibitem[{Zhou and Choi(2018)}]{zhou-choi-2018-exist}
Ethan Zhou and Jinho~D. Choi. 2018.
\newblock \href {https://www.aclweb.org/anthology/C18-1003} {They exist!
  introducing plural mentions to coreference resolution and entity linking}.
\newblock In \emph{Proceedings of the 27th International Conference on
  Computational Linguistics}, pages 24--34, Santa Fe, New Mexico, USA.
  Association for Computational Linguistics.

\end{thebibliography}
\bibliographystyle{acl_natbib}

\cleardoublepage\appendix
\section{Appendix}
\label{sec:appendix}

\subsection{Dataset}
\label{appx:dataset}

The annotation in \textit{Friends} includes plural links where a mention can belong to more than one entity clusters. We discard those mentions with plural links in our experiments and leave them as future work. All remaining mentions for \textit{Friends} are personal mentions.

\textit{BOLT} does not come with a predefined train/dev/test split. We use a random split of 80\%, 10\%, 10\% of documents in each genre for the train/dev/test split. In addition, we only use genres ``en'' and ``sm'' in \textit{BOLT}, as other genres currently do not have user IDs provided and only constitute less than 5\% documents of entire dataset. The details of our split are provided in \url{https://github.com/lxucs/online-bolt}.

\subsection{Implementation}
\label{appx:implementation}

For training on entire dialogue contexts as document input (\texttt{BL} and \texttt{SR}), we follow the similar hyperparameter settings as \citet{joshi-etal-2019-bert,spanbert-joshi,xu-choi-2020-revealing}, where long documents are split into independent segments with the maximum sequence length of 384 for SpanBERT\textsubscript{BASE}. We employ the learning rate of $2 \times 10^{-5}$ for BERT parameters and $2 \times 10^{-4}$ for task parameters with the dropout rate as $0.3$. Maximum span length is set to $6$ for \textit{Friends} and $25$ for \textit{Onto-Conv} and \textit{BOLT}. In the coarse pruning stage, we keep a maximum number of antecedents as $20$ for \textit{Friends} and $50$ for \textit{Onto-Conv} and \textit{BOLT}.

For online training and inference on the utterance sequence input (\texttt{OR}, \texttt{+SG}, \texttt{+SA}), we use one BERT segment so that the length of current utterance with past context does not exceed 384 tokens in our experiments. Gradient accumulation of 16 steps is applied during online training. We use the same learning rates and training epochs, similar as training on document input. Our best model has $\alpha_c = 1, \alpha_m = 0.1, \alpha_s = 0.1$ for the multi-task learning.

All experiments are conducted on NVIDIA TITAN RTX GPUs with 24GB memory. Training on document input takes around 3 hours and training on online input takes around 8-12 hours. All proposed methods have similar inference time, as they follow similar architecture and all operate on the online inference for prediction.

\end{document}